\documentclass[11pt,letterpaper]{article}
\usepackage{acl2016}
\usepackage{times}
\usepackage{latexsym}

\usepackage{examples}
\usepackage{balance}
\usepackage{cgloss4e}
\usepackage{verbatim}
\usepackage{cprotect}
\usepackage{listings}

\usepackage{times}
\usepackage{url}
\usepackage{latexsym}
\usepackage{amsmath}
\usepackage{stackrel}
\usepackage[pdftex]{graphicx}
\usepackage{dblfloatfix}
\usepackage{graphics}
\usepackage{array}
\newcolumntype{x}[1]{>{\centering\arraybackslash\hspace{0pt}}p{#1}}
\newcolumntype{y}[1]{<{\hspace{-.15cm}}p{#1}}
\usepackage[usenames,dvipsnames,svgnames,table]{xcolor}

\usepackage{amssymb,amsmath,epsfig}
\usepackage{mathpartir}
\usepackage{proof}
\usepackage{amsthm}
\usepackage{xspace}
\usepackage{algorithm}
\usepackage[noend]{algpseudocode}
\algrenewcommand{\algorithmicindent}{1em}

\usepackage{enumitem}

\newcommand{\notes}[1]{}

 \theoremstyle{definition}
 
\theoremstyle{plain}

\newcommand{\ith}[1]{\ensuremath{i^{{th}}}}

\newcommand{\chn}[1]{\mbox{{\it {#1}}}}

\newcount\permx
\newcount\permy
\def\permdot#1#2{
\permx=#1 \advance\permx by-1
\permy=#2 \advance\permy by-1
\psframe[fillcolor=black, fillstyle=solid]
(\permx,\permy)(#1, #2)
}

\newcommand{\boxnum}[1]{{\setlength{\fboxsep}{1pt}\raisebox{1pt}{\hspace{1pt}\fbox{\tiny #1}\hspace{1pt}}}}
\newcommand{\ind}[1]{\ensuremath{_{\kern-0.5pt\boxnum{#1}}}}

\newcommand{\yinhang}{\chn{y\'inh\'ang}\xspace}
\newcommand{\hean}{\chn{h\'e\`an}\xspace}

\def\namecite{\newcite}

\newcommand{\smallnt}[1]{\ensuremath{_{\mbox{\tiny PP}}}\xspace}

\newcommand{\pseudocode}{Algorithm}
\floatname{algorithm}{\pseudocode}

\iffalse

\else

\fi

\usepackage{multirow}
\usepackage{tikz}
\usetikzlibrary{positioning}
\usepackage{tikz-qtree}
\usetikzlibrary{arrows}

\title{Vocabulary Manipulation for Neural Machine Translation\thanks{\ \ Accepted as a short paper in ACL 2016.}}

\author{
Haitao Mi \;\;  \;\; Zhiguo Wang \;\;  \;\; Abe Ittycheriah\\
T.J.~Watson Research Center \\
IBM \\
{\tt \{hmi, zhigwang, abei\}@us.ibm.com}
}
\date{}

\begin{document}
\maketitle

\begin{abstract}
In order to capture rich language phenomena, neural machine translation models have to 
use a large vocabulary size, which requires high computing time and large memory usage.
In this paper, we alleviate this issue by introducing 
a sentence-level or batch-level vocabulary, which is only a very small sub-set of 
the full output vocabulary. For each sentence or batch, we only predict the target words in 
its sentence-level or batch-level vocabulary. Thus, we reduce both the computing time and the memory usage.
Our method simply takes into account the 
translation options of each word or phrase in the source sentence, and  
picks a very small target vocabulary
for each sentence 
based on a word-to-word translation model or a bilingual phrase library learned 
from a traditional machine translation model.
Experimental results on the large-scale English-to-French task show that our method 
achieves better translation performance by 1 BLEU point
over the large vocabulary neural machine translation system of \namecite{jean+:2015}.
\end{abstract}

\section{Introduction}
\label{sec:intro}

Neural machine translation (NMT)~\cite{bahdanau+:2014} has gained popularity in recent two years. 
But it can only handle a small vocabulary size due to the computational complexity.
In order to capture rich language phenomena and have a better word coverage,
 neural machine translation models have to 
use a large vocabulary. 

\namecite{jean+:2015} alleviated the large vocabulary issue by 
proposing an approach that 
partitions the training corpus and defines a subset of the full target vocabulary for each partition.
Thus, they only use a subset vocabulary for each partition in the training procedure 
without increasing computational complexity. 
However, there are still some drawbacks of \namecite{jean+:2015}'s method.
First, the importance sampling is simply based on the sequence of training sentences,
which is not linguistically motivated, 
thus, translation ambiguity may not be captured in the training. 
Second, the target vocabulary for each training batch is fixed 
in the whole training procedure.
Third, the target vocabulary size for each batch during training still needs to be 
as large as $30k$, so the computing time is still high. 

In this paper, we alleviate the above issues by introducing a sentence-level vocabulary,
which is very small compared with the full target vocabulary.
In order to capture the translation ambiguity, we generate those sentence-level vocabularies by
utilizing word-to-word and phrase-to-phrase translation models 
which are learned from a traditional phrase-based machine translation system (SMT).
Another motivation of this work is to combine the merits of both traditional SMT and NMT,
since training an NMT system usually takes several weeks, while the word alignment and rule extraction for SMT are much faster 
(can be done in one day). Thus, 
for each training sentence, we build a separate target vocabulary which is the 
union of following three parts:
\begin{enumerate}[topsep=0pt,itemsep=-1ex,partopsep=1ex,parsep=1ex]
\item[$\bullet$] target vocabularies of word and phrase translations 
that can be applied to the current sentence. (to capture the translation ambiguity)
\item[$\bullet$] top $2k$ most frequent target words. (to cover the unaligned target words)
\item[$\bullet$] target words in the reference of the current sentence. (to make the reference reachable)
\end{enumerate} 
As we use mini-batch in the training procedure, 
we merge the target vocabularies of all the sentences in each batch,
and update only those related parameters for each batch.
In addition, we also shuffle the training sentences at the beginning of each epoch, 
so the target vocabulary for a specific sentence varies in each epoch.
In the beam search for the development or test set, 
we apply the similar procedure for each source sentence, 
except the third bullet (as we do not have the reference) and mini-batch parts.
Experimental results on large-scale English-to-French task (Section~\ref{sec:exps}) show that our method 
achieves significant improvements 
over the large vocabulary neural machine translation system.

\section{Neural Machine Translation}
\label{sec:nmt}
\begin{figure}[!t]
\centering
\includegraphics[width=0.4\textwidth]{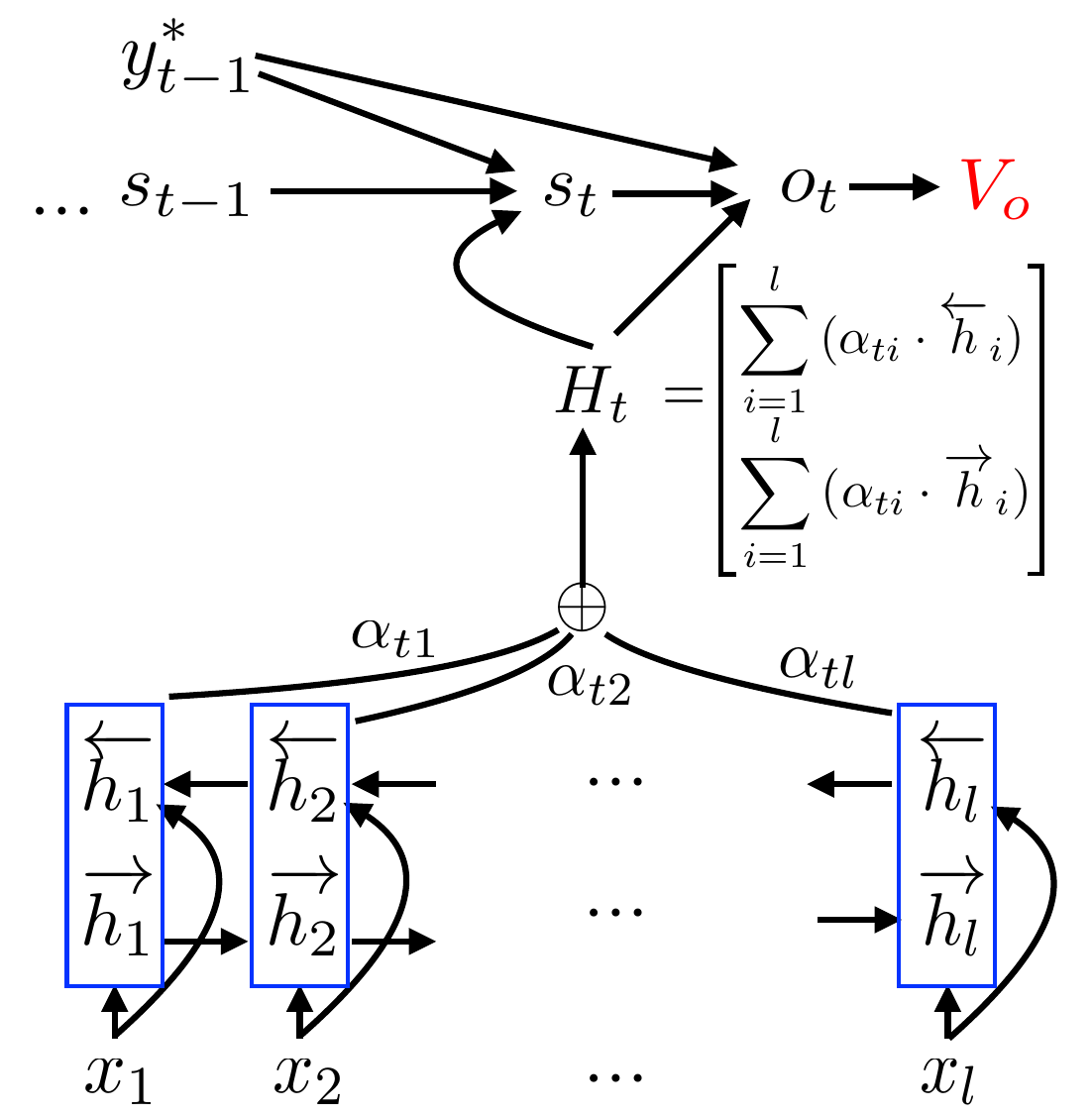}
\caption{The attention-based NMT architecture.
$\overleftarrow{h_i}$ and $\overrightarrow{h_i}$ are bi-directional encoder states.
$\alpha_{tj}$ is the attention prob at time $t$, position $j$.
$H_{t}$ is the weighted sum of encoding states.
$s_t$ is the hidden state.
$o_t$ is an intermediate output state. 
A single feedforward layer projects $o_t$ to a target vocabulary $V_o$, 
and applies softmax to predict the probability distribution over the output 
vocabulary.}
\label{fig:att}
\vspace{-0.5cm}
\end{figure}

As shown in Figure~\ref{fig:att}, neural machine translation \cite{bahdanau+:2014} 
is an encoder-decoder network. The encoder 
employs a bi-directional recurrent neural network to 
encode the source sentence ${\bf{x}}=({x_1, ... , x_l})$, 
where $l$ is the sentence length, into
a sequence of hidden states ${\bf{h}}=({h_1, ..., h_l})$,
each $h_i$ is a concatenation of a left-to-right $\overrightarrow{h_i}$
and a right-to-left $\overleftarrow{h_i}$,
\[
h_{i} = 
\begin{bmatrix}
\overleftarrow{h}_i \\ 
\overrightarrow{h}_i \\
\end{bmatrix}
=
\begin{bmatrix}
\overleftarrow{f}(x_i, \overleftarrow{h}_{i+1}) \\
\overrightarrow{f}(x_i, \overrightarrow{h}_{i-1}) \\
\end{bmatrix},
\]
where $\overleftarrow{f}$ and $\overrightarrow{f}$ 
are two gated recurrent units (GRU).

Given ${\bf h}$, the decoder predicts the target translation
by maximizing the conditional log-probability of the 
correct translation ${\bf y^*} = (y^*_1, ... y^*_m)$, where 
$m$ is the length of target sentence. At each time $t$, 
the probability of each word $y_t$ from a target vocabulary $V_y$ is:
\begin{equation}
\label{eq:py}
p(y_t|{\bf h}, y^*_{t-1}..y^*_1) \propto  \exp(g(s_t, y^*_{t-1}, H_t)),
\end{equation}
where $g$ is 
a multi layer feed-forward neural network, which takes
the embedding of the previous word $y^*_{t-1}$,
the hidden state $s_t$, and the context state $H_t$ as input. 
The output layer of $g$ is a target vocabulary $V_o$, $y_t \in V_o$ in the training procedure.
$V_o$ is originally defined as the full target vocabulary $V_y$~\cite{cho+:2014}.
We apply the softmax function over the output layer, and get the probability of 
$p(y_t|{\bf h}, y^*_{t-1}..y^*_1)$. 
In Section~\ref{sec:targetvoc}, we differentiate $V_o$ from $V_y$ by 
adding a separate and sentence-dependent $V_o$ for each source sentence.
In this way, we enable to maintain a large $V_y$, and use a small $V_o$ for each sentence.

The $s_t$ is computed as:
\begin{equation}
s_t = q(s_{t-1}, y^*_{t-1}, c_{t})
\end{equation}
\begin{equation}
c_t = 
\begin{bmatrix}
\sum_{i=1}^{l}{(\alpha_{ti} \cdot \overleftarrow{h}_i)} \\
\sum_{i=1}^{l}{(\alpha_{ti} \cdot \overrightarrow{h}_i)} \\
\end{bmatrix},
\end{equation}
where $q$ is a GRU, $c_t$ is a weighted sum of ${\bf h}$,
the weights, $\alpha$, are computed with a feed-forward neural network $r$:
\begin{equation}
\alpha_{ti} = \frac{\exp\{r(s_{t-1}, h_{i}, y^*_{t-1})\}}{\sum_{k=1}^{l}{\exp\{r(s_{t-1}, h_{k}, y^*_{t-1})\}}}
\end{equation}

\section{Target Vocabulary}
\label{sec:targetvoc}
The output of function $g$ is the probability distribution over the 
target vocabulary $V_o$.
As $V_o$ is defined as $V_y$ in~\namecite{cho+:2014},
the softmax function over $V_o$ requires 
to compute all the scores for all words in $V_o$, and results in a high computing complexity.
Thus, \namecite{bahdanau+:2014} only uses top $30k$ most frequent words for both $V_o$ and $V_y$,
and replaces all other words as {\em unknown words} (UNK).

\subsection{Target Vocabulary Manipulation}
\label{sec:dynamicvoc}
In this section, we aim to use a large vocabulary of $V_y$ (e.g. $500k$, to have a better word coverage),
and, at the same, to reduce the size of $V_o$ as small as possible (in order to reduce the computing time).
Our basic idea is to maintain a separate and small vocabulary $V_o$
for each sentence
so that we only need to compute the probability distribution of $g$ over a small vocabulary
for each sentence. 
Thus, we introduce a sentence-level vocabulary $V_{\bf{x}}$ to be our $V_o$, which depends on the sentence $\bf{x}$.
In the following part, we show how we generate the sentence-dependent $V_{\bf{x}}$.

The first objective of our method aims to capture the real translation ambiguity
for each word, and the target vocabulary of a sentence $V_o = V_{\bf{x}}$ is supposed to cover 
as many as those possible translation candidates. 
Take the English to Chinese translation for example, 
the target vocabulary for the English word $bank$ should 
contain \yinhang (a financial institution) and \hean (sloping land) in Chinese.

So we first use a word-to-word translation dictionary to generate some target vocaularies for $\bf{x}$. 
Given a dictionary $D(x) = [y_{1}, y_{2}, ...]$, where $x$ is a source word, 
$[y_{1}, y_{2}, ...]$ is a sorted list of candidate translations, 
we generate a target vocabulary $V_{\bf x}^{D}$ for a sentence ${\bf x}=(x_1, ..., x_l)$ 
by merging all the candidates of all words $x$ in $\bf{x}$. 
\[
V_{\bf x}^{D} = \bigcup_{i=1}^{l}D(x_i)
\]

As the word-to-word translation dictionary only focuses on the source words,
it can not cover the target unaligned functional or content words,
where the traditional phrases are designed for this purpose.
Thus, in addition to the word dictionary, 
given a word aligned training corpus, we also extract phrases $P(x_1 ...x_i) = [y_1, ..., y_j]$,
where $x_1 ... x_i$ is a consecutive source words, 
and $[y_1, ..., y_j]$ is a list of target words\footnote{Here we change the definition 
of a phrase in traditional SMT, 
where the $[y_1, ... y_j]$ should also be a consecutive target words. But our task in this paper 
is to get the target vocabulary, so we only care about the target word set, not the order.}.
For each sentence $\bf{x}$, we collect all the phrases that can be applied to sentence $\bf{x}$,
e.g. $x_1 ... x_i$ is a sub-sequence of sentence $\bf{x}$.
\[
V_{\bf x}^{P} = \bigcup_{\forall x_i...x_j \in {\text subseq}(\bf{x})}P(x_i...x_j),
\]
where $subseq(\bf{x})$ is all the possible sub-sequence of $\bf{x}$ with a length limit.

In order to cover target un-aligned functional words, we need 
top $n$ most common target words.
\[
V_{\bf x}^{T} = T(n).
\]

{\bf Training:}
in our training procedure, our optimization objective is to maximize the log-likelihood over the 
whole training set. In order to make the reference reachable, besides 
$V_{\bf x}^{D}$, $V_{\bf x}^{P}$ and $V_{\bf x}^{T}$,
we also need to include the target words in the reference $\bf{y}$,
\[
V_{\bf x}^{R} = \bigcup_{\forall y_i \in \bf{y}}y_i,
\]
where $\bf{x}$ and $\bf{y}$ are a translation pair.
So for each sentence $\bf{x}$, we have a target vocabulary $V_{\bf x}$:
\[
V_{\bf x} = V_{\bf x}^{D} \cup V_{\bf x}^{P} \cup V_{\bf x}^{T} \cup V_{\bf x}^{R}
\]
Then, we start our mini-batch training by randomly shuffling 
the training sentences before each epoch. 
For simplicity, we use the union of all $V_{\bf x}$ in a batch, 
\[
V_{o} = V_{\bf b} = V_{\bf x_1} \cup V_{\bf x_2} \cup ... V_{\bf x_b},
\]
where $b$ is the batch size.
This merge gives an advantage that $V_b$ changes dynamically in each epoch, 
which leads to a better coverage of parameters.

{\bf Decoding:} different from the training, 
the target vocabulary for a sentence $\bf{x}$ is 
\[
V_{o} = V_{\bf x} = V_{\bf x}^{D} \cup V_{\bf x}^{P} \cup V_{\bf x}^{T},
\]
and we do not use mini-batch in decoding.

\section{Related Work}
\label{sec:related}
To address the large vocabulary issue in NMT, 
\namecite{jean+:2015} propose a method to 
use different but small sub vocabularies for different partitions of the training corpus. 
They first partition the training set. Then, for each partition, 
they create a sub vocabulary $V_p$, and only predict and apply softmax over the 
vocabularies in $V_p$ in training procedure.
When the training moves to the next partition, 
they change the sub vocabulary set accordingly.

Noise-contrastive estimation~\cite{gutmann+:2010,mnih+:2012,mikolov+:2013,mnih+:2013} and
hierarchical classes~\cite{mnih+hinton:2009} 
are introduced to stochastically approximate the target word probability.
But, as suggested by \namecite{jean+:2015}, those methods are only designed to reduce the 
time complexity in training, not for decoding.

\begin{table*}[t]
\centering 
\tabcolsep=0.3cm
\begin{tabular}{c||c|c|c|c||c|c|c|c|c|c}
\multirow{2}{*}{set} & \multirow{2}{*}{$V_{\bf x}^P$}  & \multicolumn{3}{c||}{$V_{\bf x}^D$} & \multicolumn{3}{c|}{$V_{\bf x}^P \cup V_{\bf x}^D$} & \multicolumn{3}{c}{$V_{\bf x}^P \cup V_{\bf x}^D \cup V_{\bf x}^T$} \\
                      \cline{3-11}
                     &            & 10 & 20 & 50 & 10 & 20 & 50 & 10 & 20 & 50 \\
\hline
train  & 73.6       & 82.1 & 87.8 & 93.5 & 86.6 & 89.4 & 93.7 & \bf{92.7} & 94.2 & 96.2  \\
\hline
development   & 73.5       & 80.0 & 85.5 & 91.0 & 86.6 & 88.4 & 91.7 & \bf{91.7} & 92.7 & 94.3 \\
\hline
\end{tabular}
\caption{The average reference coverage ratios (in word-level) on the training and development sets.
We use fixed top $10$ candidates for each phrase when generating $V_{\bf x}^P$, and top $2k$ most common words for $V_{\bf x}^T$.
Then we check various top $n$ (10, 20, and 50) candidates for the word-to-word dictionary for $V_{\bf x}^D$.
\label{tab:avgcov}}
\vspace{-0.5cm}
\end{table*}

\section{Experiments}
\label{sec:exps}

\subsection{Data Preparation}
\begin{table}[t]
\centering
\tabcolsep=0.1cm
\begin{tabular}{c|c|c|c}
\multirow{2}{*}{system}   & \multicolumn{2}{c|}{train} & dev.  \\
\cline{2-4}
                       & sentence & mini-batch  & sentence \\
\hline\hline
Jean (2015) & $30k$ & $30k$ & $30k$ \\
\hline
Ours        & $2080$& $6153$& $2067$ \\
\hline
\end{tabular}
\caption{Average vocabulary size for each sentence or mini-batch (80 sentences).
The full vocabulary is $500k$, all other words are UNKs.\label{tab:avgvoc}}
\vspace{-0.5cm}
\end{table}

We run our experiments on English to French (En-Fr) task.
The training corpus consists of approximately 12 million 
sentences, which is identical to the set of \namecite{jean+:2015} and \namecite{ilya+:2015}.
Our development set is the concatenation of 
news-test-2012 and news-test-2013, which has 6003 sentences in total.
Our test set has 3003 sentences from WMT news-test 2014.
We evaluate the translation quality using the case-sensitive BLEU-4 metric~\cite{BLEU:2002}
with the multi-bleu.perl 
script.

Same as \namecite{jean+:2015}, our full vocabulary size 
is $500k$,
we use AdaDelta~\cite{adadelta}, 
and mini-batch size is 80. 
Given the training set, we first run the `fast\_align'~\cite{dyer+:2013} in one direction, 
and use the translation table as our word-to-word dictionary.
Then we run the reverse direction and apply `grow-diag-final-and' heuristics to get the alignment.
The phrase table is extracted with a standard algorithm in Moses~\cite{koehn+:2007}. 

In the decoding procedure, our method is very similar to the `candidate list' of \namecite{jean+:2015}, 
except that we also use bilingual phrases and we only include top $2k$ most frequent target words.
Following \namecite{jean+:2015}, we dump the alignments for each sentence, and 
replace UNKs with the word-to-word dictionary or the source word.

\subsection{Results}

\subsubsection{Reference Reachability}
The reference coverage or reachability ratio is very important when we limit the target vocabulary 
for each source sentence,
since we do not have the reference in the decoding time, 
and we do not want to narrow the search space into a bad space.
Table~\ref{tab:avgcov} shows the average reference coverage ratios (in word-level) 
on the training and development sets.
For each source sentence $\bf{x}$, $V_{\bf{x}}^*$ here is 
a set of target word indexes (the vocabulary size is $500k$, others are mapped to UNK).
The average reference vocabulary size $V_{\bf{x}}^R$ for each sentence is 23.7 on the training set (22.6 on the dev. set).
The word-to-word dictionary $V_{\bf{x}}^D$ has a better coverage than phrases $V_{\bf{x}}^P$,
and when we combine the three sets we can get better coverage ratios.
Those statistics suggest that we can not use each of them alone due to the low reference coverage ratios.
The last three columns show three combinations, all of which have higher than 90\% coverage ratios.
As there are many combinations, training an NMT system is time consuming, 
and we also want to keep the output vocabulary size small 
(the setting in the last column in Table~\ref{tab:avgcov} results in an average $11k$ vocabulary size for mini-batch 80), 
thus, in the following part, 
we only run one combination (top 10 candidates for both $V_{\bf{x}}^P$ and $V_{\bf{x}}^D$, 
and top $2k$ for $V_{\bf{x}}^T$),
where the full sentence coverage ratio is 20.7\% on the development set.

\begin{table*}[t]
\centering
\tabcolsep=0.2cm
\begin{tabular}{c|c|c|c|c|c|c}
top $n$ common words & 50 & 200 & 500 & 1000 & 2000 & 10000 \\
\hline
BLEU on dev.         & 30.61 & 30.65 & 30.70 & 30.70 & 30.72 & 30.69 \\
\hline
avg. size of $V_o = V_{\bf x}^P \cup V_{\bf x}^D \cup V_{\bf x}^T$ & 202 & 324 & 605 & 1089 & 2067  &  10029 \\
\hline
\end{tabular}
\caption{Given a trained NMT model, we decode the development set with various top $n$ most common target words.
For En-Fr task, the results suggest that we can reduce the $n$ to 50 without losing much in terms of BLEU score.
The average size of $V_o$ is reduced to as small as 202, which is significant lower than 2067 
(the default setting we use in our training). 
\label{tab:vark}}
\vspace{-0.5cm}
\end{table*}

\subsubsection{Average Size of $V_o$}
With the setting shown in {\bf bold} column in Table~\ref{tab:avgcov}, 
we list average vocabulary size of \namecite{jean+:2015} and ours in Table~\ref{tab:avgvoc}.
\namecite{jean+:2015} fix the vocabulary size to $30k$ for each sentence and mini-batch,
while our approach reduces the vocabulary size to 2080 for each sentence, and 6153 for each mini-batch.
Especially in the decoding time, our vocabulary size for each sentence is about 14.5 times smaller
than $30k$.

\subsubsection{Translation Results}
\begin{figure}
\includegraphics[width=0.33\textwidth,angle=-90]{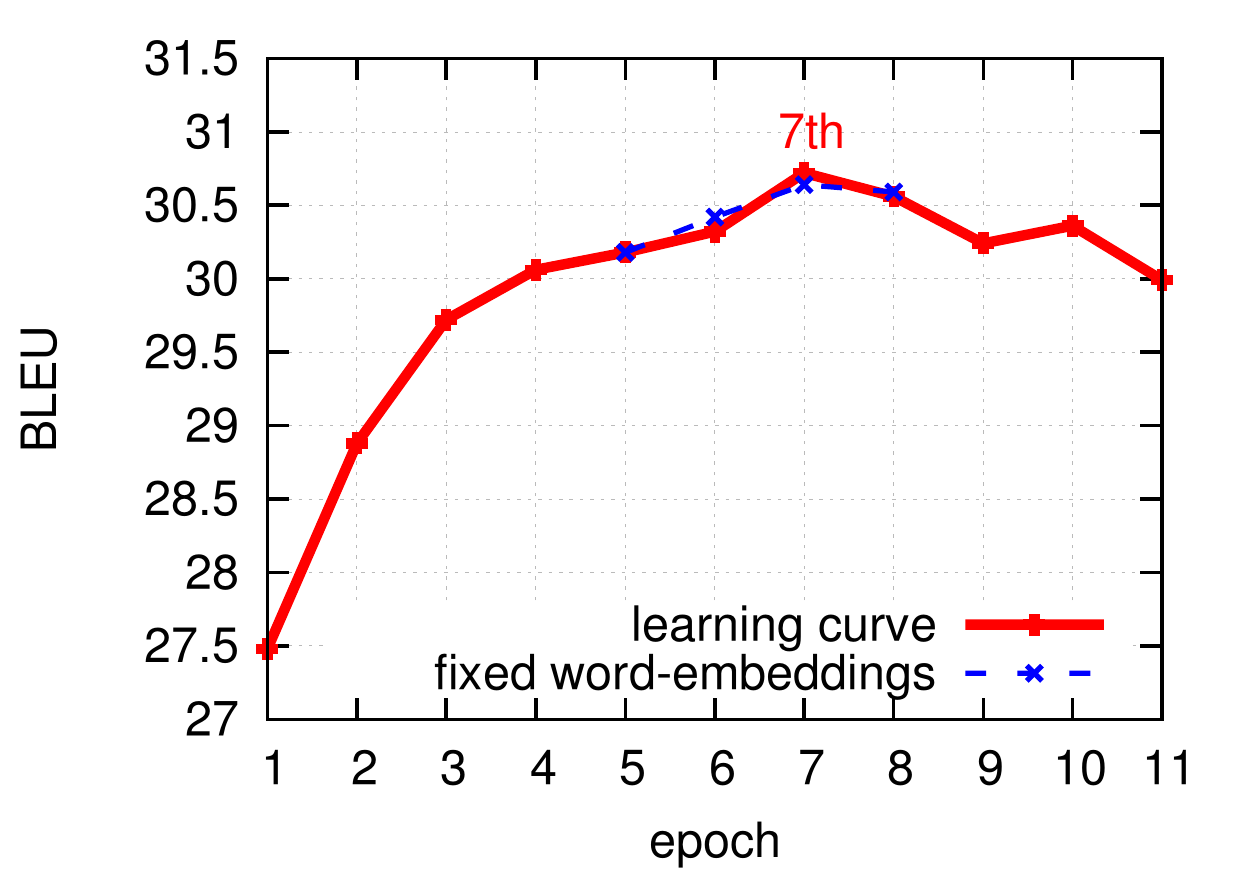}
\caption{The learning curve on the development set.
An epoch means a complete update through the full training set.
\label{fig:learningc}}
\vspace{-0.3cm}
\end{figure}

The red solid line in Figure~\ref{fig:learningc}
shows the learning curve of our method on the 
development set, 
which picks at epoch 7 with a BLEU score of 30.72.
We also fix word embeddings at epoch 5, and 
continue several more epochs. The corresponding blue dashed line 
suggests that there is no significant difference between them. 

We also run two more experiments: 
$V_{\bf x}^D \cup V_{\bf x}^T$ and 
$V_{\bf x}^P \cup V_{\bf x}^T$ separately (always have $V_{\bf x}^R$ in training).
The final results on the test set are 34.20 and 34.23 separately.
Those results suggest that we should use both the translation dictionary and phrases 
in order to get better translation quality.

\begin{table}[t]
\centering
\tabcolsep=0.15cm
\begin{tabular}{c|c||c|c}
\multicolumn{2}{c||}{single system}   & dev.  & test \\
\hline
\multicolumn{2}{c||}{Moses from \namecite{cho+:2014}}    & N/A & 33.30 \\
\hline
\multirow{2}{*}{Jean (2015)} & candidate list            & 29.32 & 33.36 \\
                                       & +UNK replace    & 29.98 & 34.11 \\
\hline
\multirow{2}{*}{Ours}                  & voc. manipulation & 30.15 & 34.45 \\
                                       & +UNK replace    & 30.72 & 35.11 \\
\hline
\multicolumn{2}{c||}{best from \namecite{durrani+:2014}} & N/A   & 37.03 \\
\hline
\end{tabular}
\caption{Single system results on En-Fr task.
\label{tab:enfr}}
\vspace{-0.5cm}
\end{table}

Table~\ref{tab:enfr} shows the single system results on En-Fr task.
The standard Moses in~\namecite{cho+:2014} on the test set is 33.3.
Our target vocabulary manipulation achieves a BLEU score of 34.45 on the test set, 
and 35.11 after the UNK replacement. 
Our approach improves the translation quality by 1.0 BLEU point on the test set 
over the method of \namecite{jean+:2015}. 
But our single system is still about 2 points behind of the best phrase-based system~\cite{durrani+:2014}.

\subsubsection{Decoding with Different Top $n$ Most Common Target Words }
Another interesting question is what is the performance if we vary the size top $n$ most common target words
in $V_{\bf{x}}^T$. As the training for NMT is time consuming, 
we vary the size $n$ only in the decoding time. Table~\ref{tab:vark} shows the BLEU scores on the development set.
When we reduce the $n$ from 2000 to 50, we only loss 0.1 points, 
and the average size of sentence level $V_o$ is reduced to 202, which is significant smaller than 2067 (shown in Table~\ref{tab:avgvoc}).
But we should notice that we train our NMT model in the condition of the {\bf bold} column in Table~\ref{tab:avgvoc}, 
and only test different $n$ in our decoding procedure only. 
Thus there is a mismatch between the training and testing when $n$ is not 2000.

\subsubsection{Speed}
In terms of speed, as we have different code bases\footnote{
Two code bases share the same architecture, initial states, and hyper-parameters.
We simulate \namecite{jean+:2015}'s work with 
our code base in the both training and test procedures, the final results of our simulation 
are 29.99 and 34.16 on dev. and test sets respectively. Those scores are very close to
\namecite{jean+:2015}.} between \namecite{jean+:2015} and us,
it is hard to conduct an apple to apple comparison.
So, for simplicity, we run another experiment with our code base, 
and increase $V_{\bf b}$ size to $30k$ for each batch (the same size in \namecite{jean+:2015}). 
Results show that increasing the $V_{\bf b}$ to $30k$ slows down the training speed by 1.5 times.

\section{Conclusion}
In this paper, we address the large vocabulary issue in neural machine translation
by proposing to use a sentence-level target vocabulary $V_o$, 
which is much smaller than the full target vocabulary $V_y$. 
The small size of $V_o$ reduces the computing time of the softmax function in each predict step,
while the large vocabulary of $V_y$ enable us to model rich language phenomena.
The sentence-level vocabulary $V_o$ is generated with the traditional word-to-word and phrase-to-phrase translation libraries.
In this way, we decrease the size of output vocabulary $V_o$ under $3k$ for each sentence,
and we speedup and improve the large-vocabulary NMT system.

\section*{Acknowledgment}
We thank the anonymous reviewers for their comments.

\balance
\bibliographystyle{acl2016}
\bibliography{thesis}

\end{document}